\documentclass{article}
\usepackage{spconf}

\usepackage{cite}
\usepackage{amsmath,amssymb,amsfonts}
\usepackage{algorithmic}
\usepackage{graphicx}
\usepackage{textcomp}
\usepackage{xcolor}

\usepackage{mathtools}
\usepackage{tcolorbox}
\usepackage{color}
\usepackage{theorem}
\usepackage{amssymb}
\usepackage{caption}
\usepackage{subcaption}
\usepackage{cite,hyperref}
\usepackage{cases}
\usepackage{url}
\usepackage{algorithmic}
\usepackage{algorithm}
\usepackage{tikz}
\usetikzlibrary{shapes,arrows}
\input{mysymbol.sty}
\input{my_beamer_defs.sty}
\tikzstyle{phantom vertex} = [ ellipse, 
                               anchor = center, 
                               minimum height = 1*\unit, 
                               minimum width  = 1*\unit,
                               inner sep=0pt,
                               anchor=center]
\tikzstyle{red vertex}   = [black, fill = red!20,   phantom vertex, draw]
\tikzstyle{black vertex} = [black, fill = black!20, phantom vertex, draw]
\tikzstyle{blue vertex}  = [black, fill = blue!20,  phantom vertex, draw]
\tikzstyle{green vertex} = [black, fill = green!20,  phantom vertex, draw]
\tikzstyle{yellow vertex} = [black, fill = yellow!20,  phantom vertex, draw]
\tikzstyle{cyan vertex} = [black, fill = cyan!20,  phantom vertex, draw]
\tikzstyle{vertex}       = [draw, phantom vertex]

\tikzstyle{point} = [ellipse, inner sep=0pt, draw, fill=white, anchor = center,
                     minimum height = 0.05*\unit, minimum width  = 0.05*\unit]

\newtcolorbox{myblockt}[1]{colback=urblue!5!white,
	colframe=urblue,fonttitle=\bfseries,
	title=#1}

\newtcolorbox{myblock}{colback=urblue!5!white,
	colframe=urblue,fonttitle=\bfseries}

\def\BibTeX{{\rm B\kern-.05em{\sc i\kern-.025em b}\kern-.08em
    T\kern-.1667em\lower.7ex\hbox{E}\kern-.125emX}}

\begin{document}
\ninept

\title{Graphon-aided Joint Estimation of Multiple Graphs}
\name{Madeline Navarro and Santiago Segarra
\thanks{ This work was partially supported by NSF under award CCF-2008555. 
	Emails:  \href{mailto:nav@rice.edu}{nav@rice.edu}, \href{mailto:segarra@rice.edu}{segarra@rice.edu}}}
\address{Electrical and Computer Engineering, Rice University, USA}

\maketitle

\begin{abstract}%
We consider the problem of estimating the topology of multiple networks from nodal observations, where these networks are assumed to be drawn from the same (unknown) random graph model.
We adopt a graphon as our random graph model, which is a nonparametric model from which graphs of potentially different sizes can be drawn.
The versatility of graphons allows us to tackle the joint inference problem even for the cases where the graphs to be recovered contain different number of nodes and lack precise alignment across the graphs.
Our solution is based on combining a maximum likelihood penalty with graphon estimation schemes and can be used to augment existing network inference methods. 
We validate our proposed approach by comparing its performance against competing methods in synthetic and real-world datasets.
\end{abstract}


\section{Introduction}

Networks (or graphs) are powerful representations of complex information due to their ability to represent structure via dyadic relationships.
Many fields of research utilize network structures for representing and analyzing complex data, such as ecology for predicting animal behavior \cite{farine2015constructing}, neuroscience for modeling relationships between neurons \cite{narayan2016mixed}, and environmental science for discovering and predicting outcomes of climate relationships \cite{donges2009complex}.

While networks are convenient and interpretable tools for tasks on complex data, knowledge of the underlying structure may be unavailable, as is the case for functional connectivity between brain regions \cite{narayan2016mixed}, or the underlying network may be expensive to obtain, as with structural (anatomical) connectivity between neurons \cite{sporns2013human}. 
The inference of network connectivity from nodal observations is a ubiquitous problem that has been well studied in fields such as statistics \cite{kolaczyk2009statistical} and signal processing \cite{mateos2019connecting}.
Data-driven methods for the inference of network structure include graphical models \cite{friedman2008sparse,meinshausen2006high}, structural equation models \cite{cai2013sparse}, and graph signal processing-based approaches \cite{mateos2019connecting,kalofolias2016learn,segarra2017network}. 

In many of the above mentioned applications, it is often more important to infer the topology of multiple networks.
For example, brain functional connectivity is a valuable tool for diagnosis, and the acquisition of multiple functional networks is necessary when considering multiple patients or scenarios \cite{narayan2016mixed}.
Additionally, a prominent scenario requiring knowledge of multiple networks is when networks vary over time.
An ecological example includes the estimation of evolving social networks for a species of interest \cite{de2011dynamics}.

In the case of inferring the topologies of multiple networks, separate estimation is a feasible methodology.
However, in many scenarios a joint inference method may achieve better performance by leveraging common structures between the graphs to be inferred.
For instance, one would expect certain levels of similarities between the brain networks of different healthy individuals or between the same social network observed at different points in time.
Prominent methods for multiple network inference include statistical approaches, primarily consisting of the joint estimation of Gaussian graphical models \cite{wang2020high,cai2016joint,gan2019bayesian,honorio2010multi,danaher2014joint}.
These methods typically involve modifications on the graphical lasso formulation with additional encouragement of structural similarity.
Estimation of time-varying graphs is widely popular, as the relationship between graphs is typically straightforward to implement by considering that graph variation is smooth across time \cite{kalofolias2017learning,yamada2019time}. 
The above methods for estimating multiple networks typically enforce similar structure, such as promoting similar sparsity patterns \cite{navarro2020joint}.

We consider the problem of {\it estimating the topology of multiple networks sampled from the same (unknown) random graph model}, where graphs have similar global structural characteristics inherited from the model.
As our (nonparametric) network model we adopt a graphon~\cite{lovasz2012large}, but we do not assume the specific graphon model to be known a priori~\cite{roddenberry2021graphon}.
While estimation of multiple networks is well-studied, to the best of our knowledge no previous method utilizes a shared graphon relationship to jointly estimate graphs of potentially different sizes. 

\vspace{1mm}
\noindent
{\bf Contributions.}
The contributions of our paper are threefold:\\
1)~We present a methodology to {\it infer multiple networks that potentially lack node alignment and may have different sizes} by leveraging the assumption that graphs come from the same nonparametric network model. \\
2)~We detail how this methodology can be combined with existing network inference methods, effectively providing a whole family of methods to solve the problem of interest. \\
3)~Through numerical experiments in synthetic and real-world data we demonstrate the performance of our method in comparison with separate inference and competing joint inference algorithms.

\section{Preliminaries}

\noindent{\bf Graph signal processing.}
We consider undirected, unweighted graphs of the form $\ccalG=(\ccalV,\ccalE)$ with node (vertex) set $\ccalV$ of cardinality $N$ and edge set $\ccalE\subseteq\ccalV\times\ccalV$. 
The structure of a graph can be represented by its graph shift operator (GSO) \cite{sandryhaila2013discrete,segarra2017optimal} as the matrix $\bbS\in\{0,1\}^{N\times N}$, where $S_{ij}\neq0$ if and only if the edge $(i,j)$ exists in the network, and $S_{ij}=0$ otherwise. 
We define graph signals as real-valued observations at each of the $N$ nodes, represented by a vector $\bbx\in\mathbb{R}^N$.
We may associate these nodal values with the graph topology via graph signal models.
Choices for graph signal models include stationary signals that result from diffusion processes over the graph~\cite{sandryhaila2013discrete,marques2017stationary} or as multivariate random numbers, where the graph structure represents statistical dependencies between variables \cite{friedman2008sparse,meinshausen2006high}.

\begin{figure*}
\centering
	\begin{minipage}[c]{.45\textwidth}
		\includegraphics[width=\textwidth]{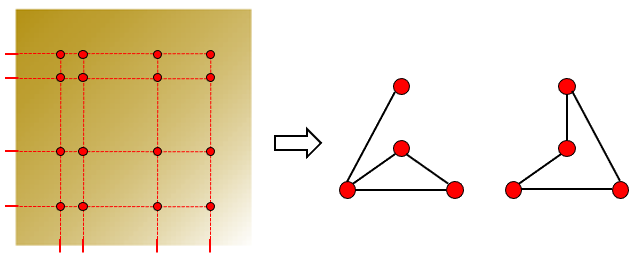}
		
		\centering{\small (a)}
	\end{minipage}
	\hspace{1.5cm}
	\begin{minipage}[c]{.45\textwidth}
		\includegraphics[width=\textwidth]{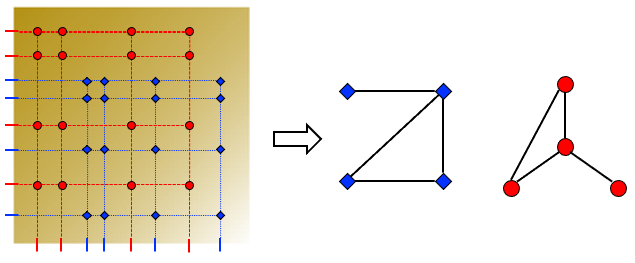}
		
		\centering{\small (b)}
	\end{minipage}
	\vspace{-2mm}
\caption{Schematic depiction of the two problem modalities considered. (a)~Multiple graphs sampled from the {\em same latent point sets} in the {\em same graphon}. Sampled graphs have not only the same size but also node alignment; see Section~\ref{Ss:prob1_ver1}. (b)~Multiple graphs sampled from {\em different latent point sets} in the {\em same graphon}. Sampled graphs may have different sizes; see Section~\ref{Ss:prob1_ver2}.}
	\vspace{-4mm}
\label{fig_cases}
\end{figure*}

\vspace{.1in}
\noindent{\bf Graphons.}
A graphon is a bounded symmetric measurable function $W: [0,1]^2\rightarrow[0,1]$ whose domain can be interpreted as edges in an infinitely large adjacency matrix, while the range of $W$ represents edge probabilities.
By this definition, a graphon can be seen as a random graph model from which graphs with similar structural characteristics can be sampled~\cite{diaconis2007graph,lovasz2012large,avella2020graphon}.
Generating an undirected graph $\ccalG=(\ccalV,\ccalE)$ from a graphon $W$ consists of two steps: (1)~selecting a random value between 0 and 1 for each node, and (2)~assigning an edge between nodes with probability equal to the value of the graphon at the their randomly sampled points. 
Formally, the steps are as follows
\begin{subequations}
\begin{alignat}{2}
&\zeta_{i} \sim \text{Uniform}([0,1]) &&\forall~i\in\ccalV,
\label{equ_graphon_samp1}\\
& S_{ij} = S_{ji} \sim \text{Bernoulli}\left( W(\zeta_i,\zeta_j) \right) &\qquad& \forall~(i,j)\in\ccalV\times\ccalV,
\label{equ_graphon_samp2}
\end{alignat}
\end{subequations}
where the latent variables $\zeta_i\in[0,1]$ are independently drawn for each node $i$.
This notion of graphon encompasses many commonly used exchangeable distributions on networks.
Indeed, Erd\H{o}s-R\'enyi graph models are represented via constant graphons~\cite{erdHos1959renyi} and stochastic block models via piecewise-constant graphons~\cite{holland1983stochastic}.

In our case, we assume that graphs are sampled from the same graphon, which is also unknown.
Therefore, we propose a method to jointly estimate both the graphs and the underlying graphon.
Estimating a graphon from observed adjacency matrices is a well-studied task \cite{gao2015rate,klopp2019optimal}, and methods to infer a graphon from binary graphs include estimating the graphon function as a continuous smooth object \cite{sischka2022based,chan2014consistent} along with the coarser stochastic blockmodel estimation \cite{olhede2014network,airoldi2013stochastic}.
Other methods provide only the estimation of the graphon points $W(\zeta_i,\zeta_j)$ where it was sampled, which is equivalent to estimating a probability matrix~\cite{chatterjee2015matrix}.

\section{Problem Statement}
Consider a set of $K$ different graphs $\{\ccalG^{(k)}\}_{k=1}^K$ where the $k$-th graph has $N^{(k)}$ nodes. 
The set of undirected, unweighted adjacency matrices is represented by the set of GSOs $\bbS=\{\bbS^{(k)}\}_{k=1}^K$.
Assume also that there is a set of graph signals provided for each graph, represented by $\bbX^{(k)}:=[\bbx_1^{(k)}~\cdots~\bbx_{r_k}^{(k)}]\in\mathbb{R}^{N^{(k)}\times r_k}$, where the $r_k$ columns contain the graph signals corresponding to the $k$-th graph.
We further assume that all graphs are sampled from the same generative model, a graphon $W$. 
We present our problem as follows.

\vspace{.1in}
\noindent{\bf Problem 1} {\it Given sets of observations $\bbX=\{\bbX^{(k)}\}_{k=1}^K$ for $K$ graphs, find the adjacency matrices $\bbS=\{\bbS^{(k)}\}_{k=1}^K$ under the assumptions that (AS1) all graphs are sampled from the same (unknown) graphon $W$ and (AS2) the latent point sets $\zeta^{(k)}$ in \eqref{equ_graphon_samp1} for each graph are known.
}
\vspace{.1in}

The first assumption (AS1) creates a relationship among the graphs, and with it we may improve estimation of graphs by jointly inferring the graph structures given their shared relationship.
The second assumption (AS2) eliminates the identifiability problem for graphon estimation, where multiple graphons can lead to the same random graph distribution \cite{diaconis2007graph}.
When all latent point sets are equivalent, i.e., $\zeta^{(k)} = \zeta$ for all $k\in\{1,2,\dots,K\}$, (AS2) is equivalent to the assumption in previous joint network inference methods, where node alignment is present and known for all pairs of graphs. 
However, assuming possibly different known latent point sets is a weaker assumption than that of previous methods, as we do not require node alignment for the graphs.

The assumption (AS2) corresponds intuitively to situations of known sensor placement, such as known locations of electrode placement for neural response data collection or known climate regions to be observed.
For example, the brain functional networks of multiple subjects may be measured by considering the same known brain regions or neurons across subjects \cite{narayan2016mixed}.
Inferred graphs may also correspond to statistical interdependence between pairs of variables in a climate data set, where variables are measured at known spatial regions of earth \cite{donges2009complex}.

\section{Graphon-aided Joint Network Estimation}

In Sections~\ref{Ss:prob1_ver1} and~\ref{Ss:prob1_ver2} we tackle two versions of Problem 1 of increasing difficulty whereas in Section~\ref{Ss:combining_methods} we explain how these solutions can be combined with existing network inference methods.

\subsection{Graphs and Probability Matrix Estimation}
\label{Ss:prob1_ver1}

First consider the case where all graphs are sampled as in \eqref{equ_graphon_samp1} from the same points in the graphon space, that is, $\zeta^{(k)}=\zeta$ for all graphs $k\in\{1,2,\dots,K\}$; see Fig.~\ref{fig_cases}a.
In practice, this case arises, e.g., when using the same sensor placement under multiple trials or experiments.
Since we only consider edge probabilities in the graphon at points $(x,y)\in\zeta\times\zeta$, we need not consider the whole graphon $W$ but only the probability matrix $\bbT\in[0,1]^{N\times N}$ that contains the edge probabilities at the sampled points.
The graphs $\bbS^{(k)}$ are then sampled from the same probability matrix $\bbT$, so the graphs all must have the same size, that is, $N^{(k)}=N$ for all $k\in\{1,2,\dots,K\}$.

Given $\bbT$, the log-likelihood of a graph $\bbS^{(k)}$ is
\begin{alignat}{2}
&\log\Pr{\bbS^{(k)}|\bbT} = \sum_{i<j} S^{(k)}_{ij}\log(T_{ij}) + (1-S^{(k)}_{ij})\log(1-T_{ij}),
\nonumber
\end{alignat}
where we have leveraged the fact that, given $\bbT$, edges are drawn independently in our graph model [cf.~\eqref{equ_graphon_samp2}].
Furthermore, we can estimate each edge probability $T_{ij}$ by the sample mean of the graph edges. 
Thus, we estimate the probability matrix $\bbT$ as $\frac{1}{K}\sum_{k=1}^K \bbS^{(k)}$.

Recalling the notation for $\bbX$ and $\bbS$ from Problem 1, consider a generic optimization problem to estimate multiple networks that we formalize as
\begin{alignat}{2}
&\min_{\bbS}~~ && f(\bbS,\bbX) + L(\bbS),
\label{equ_netinf}
\end{alignat}
where the objective function $f(\bbS,\bbX)$ estimates graph structures from the observed datasets, and $L(\bbS)$ is an additional graph penalty or regularizer; in Section~\ref{Ss:combining_methods} we provide common examples for these functions.
To solve our problem at hand, we propose to append the generic formulation in~\eqref{equ_netinf} with a negative log-likelihood penalty to obtain
\begin{alignat}{2}
&\min_{\bbS,\bbT}~~~ &~~~& f(\bbS,\bbX) + L(\bbS) - \sum_{k=1}^K \log\Pr{\bbS^{(k)}|\bbT} 
\nonumber\\
&\text{~~~s.to} && 
\bbS^{(k)}\in\ccalS_A,~~\bbT=\frac{1}{K}\sum_{k=1}^K \bbS^{(k)},
\label{equ_netinf_probmat}
\end{alignat}
where we jointly estimate the graphs and their shared generative probability matrix $\bbT$. 
The estimation of the probability matrix entries is included as the sample mean of the edges in the graphs. 
The set $\ccalS_A$ enforces valid binary adjacency matrices, that is,
\be
\ccalS_A = \left\{ \bbS : \bbS=\bbS^\top, S_{ii}=0, S_{ij}\in\{0,1\} \right\},
\ee
where we consider undirected graphs without self-loops and edges that are unweighted.

As mentioned in the problem statement, the assumption $\zeta^{(k)}=\zeta$ for all $k\in\{1,2,\dots,K\}$ is equivalent to node alignment for all graphs.
We relate \eqref{equ_netinf_probmat} to the task of estimating functional networks among the same brain regions of one subject under a set of discrete stimuli, or observing climate variables among the same geographical regions over several time instances.

\begin{figure*}
\centering
	\begin{minipage}[c]{.33\textwidth}
		\includegraphics[width=1.1\textwidth]{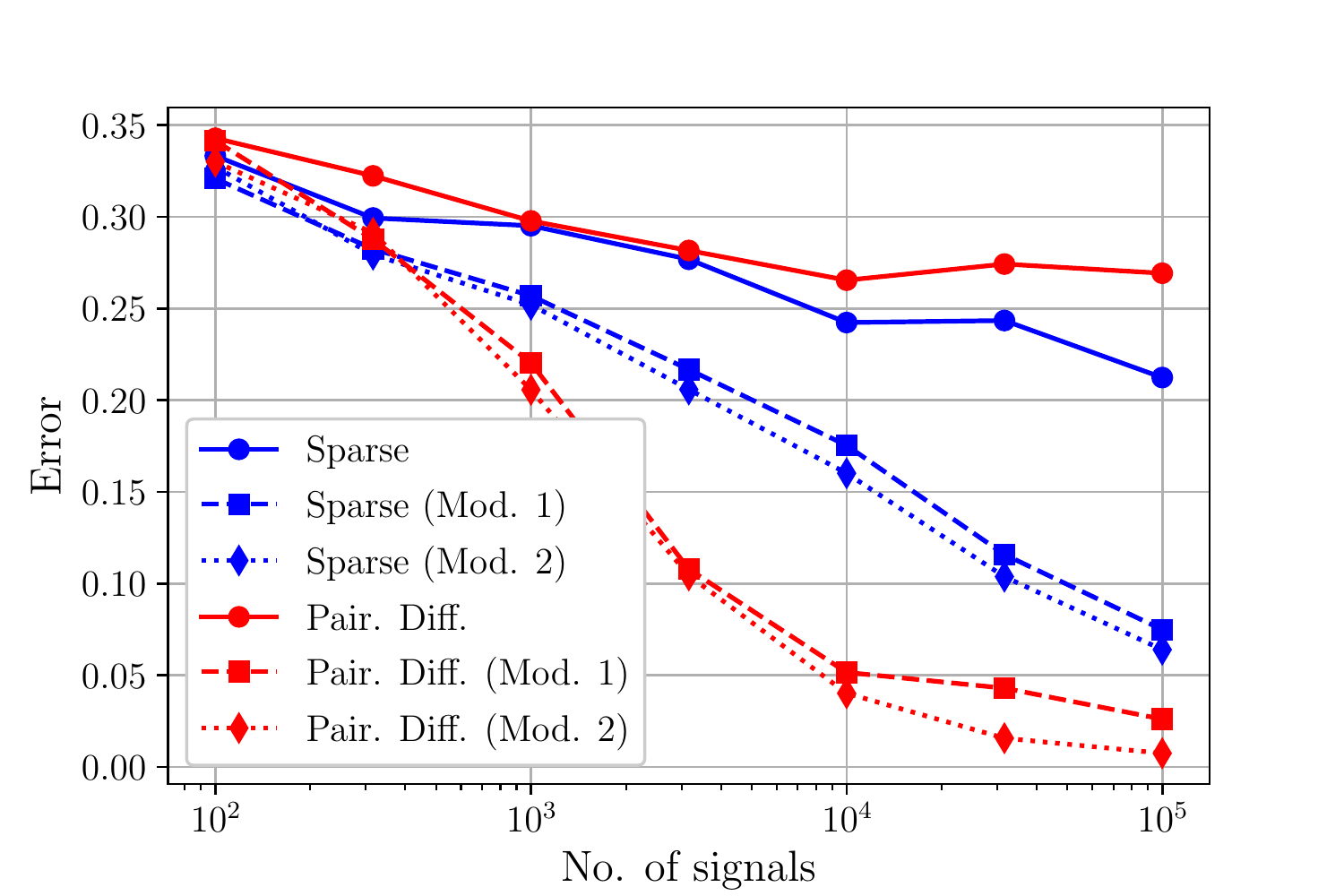}
		
		\centering{\small (a)}
	\end{minipage}
	\begin{minipage}[c]{.33\textwidth}
		\includegraphics[width=1.1\textwidth]{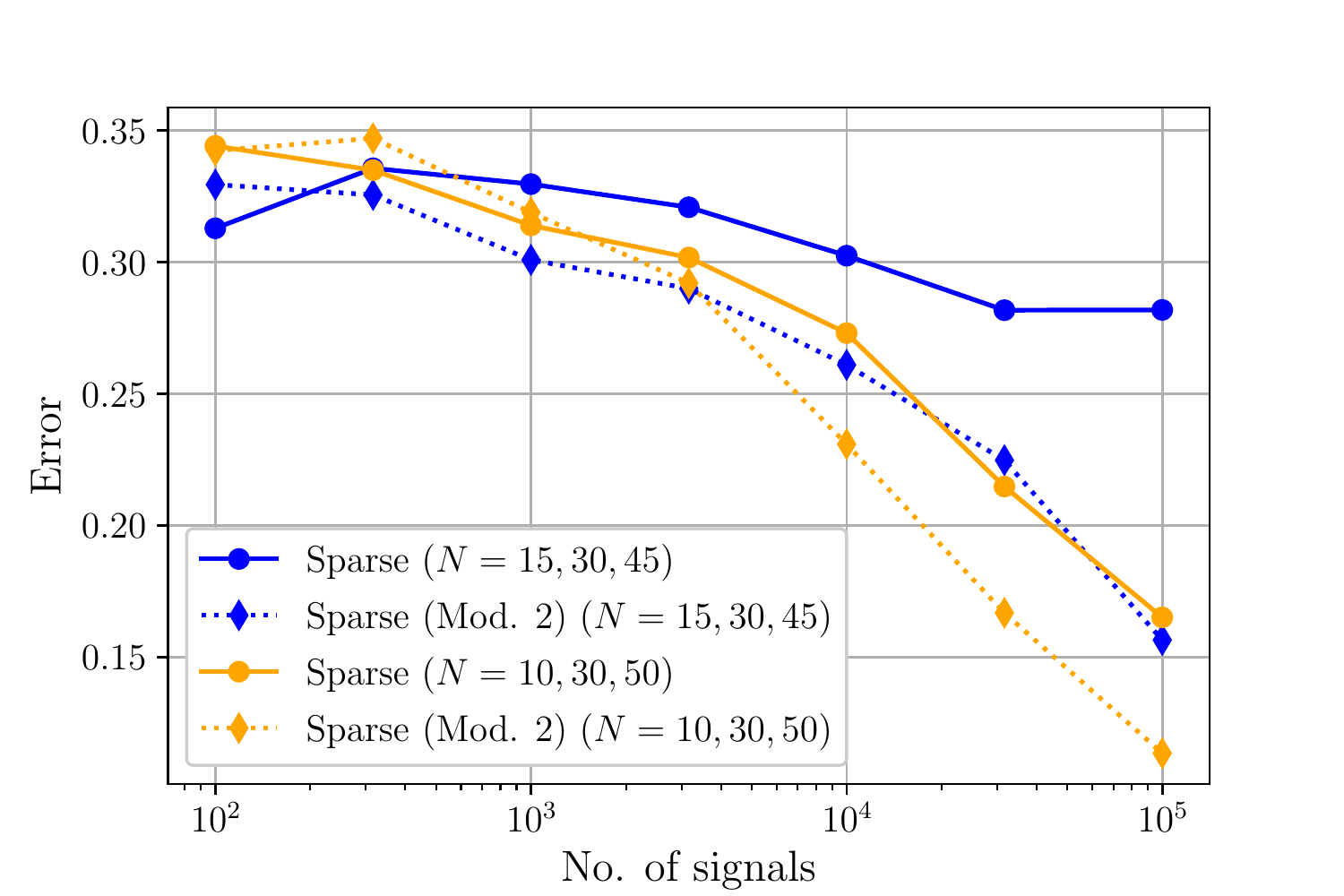}
		
		\centering{\small (b)}
	\end{minipage}
	\begin{minipage}[c]{.33\textwidth}
		\includegraphics[width=1.1\textwidth]{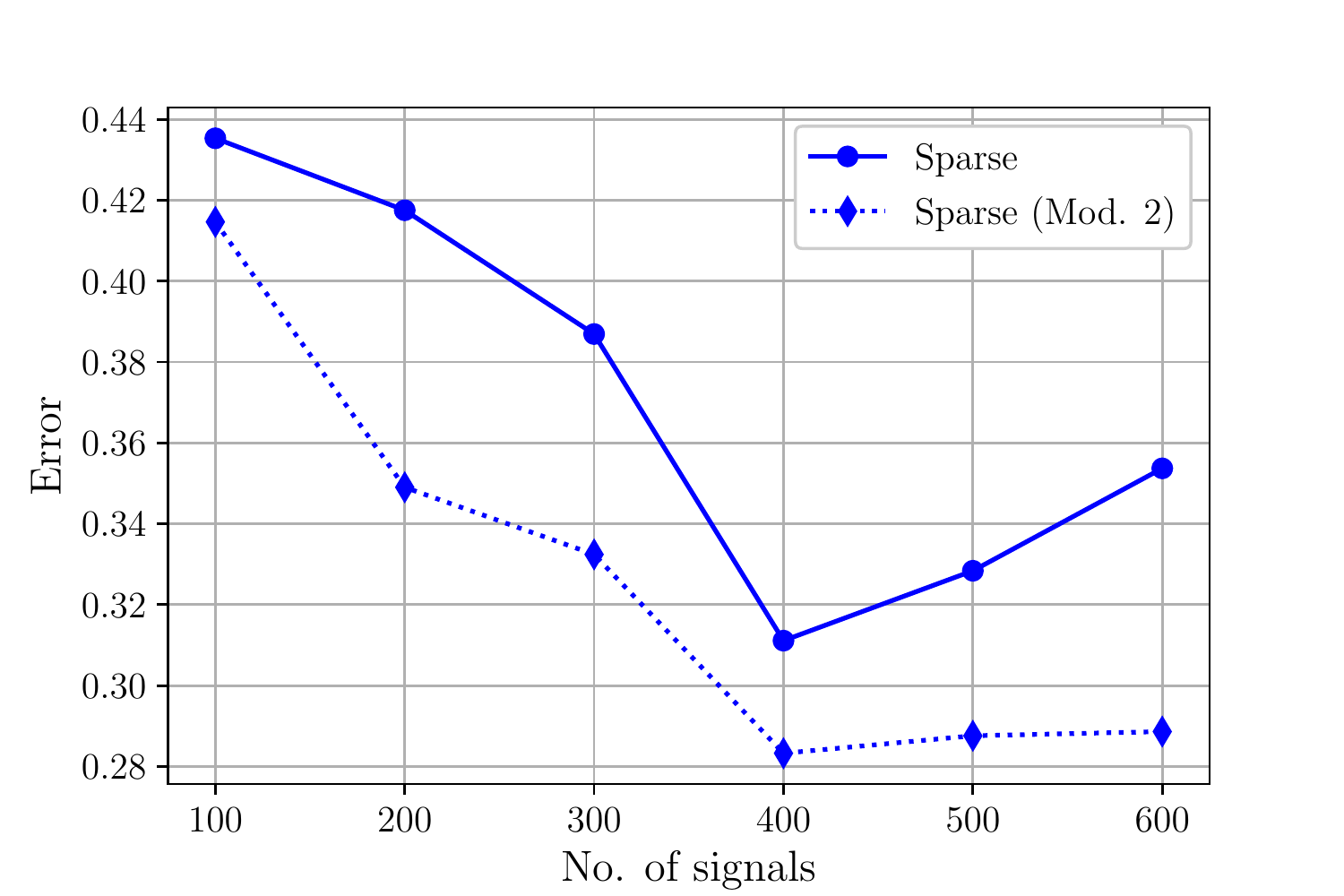}

		\centering{\small (c)}
	\end{minipage}
\caption{(a)~Recovery error for $K=3$ graphs sampled from the {\it same latent point sets} in the same graphon as a function of the number of observed signals. Incorporating the joint estimation of the probability matrix or the graphon both improve estimation performance.~(b)~Recovery error for $K=3$ graphs sampled from {\it latent point sets of different sizes} in the same graphon as a function of the number of observed signals. Separate inference of graphs is outperformed by including joint estimation of the underlying graphon.~(c)~Recovery error of {\it induced subgraphs} of three senate networks of sizes $N=15,30,45$ as a function of the number of observed signals. Joint network and graphon inference outperforms separate network inference for all sets of observed signals.}
\label{fig_exp}
\end{figure*}

\subsection{Graphs and Graphon Estimation}
\label{Ss:prob1_ver2}

We now consider the case where each graph is sampled from different latent point sets $\zeta^{(k)}$, and graphs may possibly have different sizes $N^{(k)}$; see~Fig. \ref{fig_cases}b.
Therefore, each graph is sampled from a potentially different probability matrix $\bbT^{(k)}$, which is the value of the graphon at the points $(x,y)\in\zeta^{(k)}\times\zeta^{(k)}$.
The probability matrices provide estimates of the graphon at the known latent point pairs, and each graph provides information about the value of its respective probability matrix.
We present an optimization problem to jointly estimate the graphs, the probability matrices, and the graphon as 
\begin{alignat}{2}
&\min_{\scriptsize
\bbS,\bbT,W
} &~~~& 
f(\bbS,\bbX) + L(\bbS) - \sum_{k=1}^K \log\Pr{\bbS^{(k)}|\bbT^{(k)}} + g(W)
\nonumber\\
&\text{~~~~s.to} && \bbS^{(k)}\in\ccalS_A,
\nonumber\\
& &&\bbT^{(k)}=h(\bbS^{(k)}),~~~T^{(k)}_{ij} = W(\zeta_i^{(k)},\zeta_j^{(k)}),
\nonumber\\
& && W:[0,1]^2\rightarrow [0,1], ~~~W(x,y)=W(y,x),
\label{equ_netinf_graphon}
\end{alignat}
where we include the same negative log-likelihood term as in~\eqref{equ_netinf_probmat}, but each graph is associated with a different probability matrix $\bbT^{(k)}$. 
The function $h(\cdot)$ is a probability matrix estimation method that takes an adjacency matrix as input, such as network histogram or stochastic block model approximations \cite{olhede2014network,chan2014consistent}.
The third constraint fits the graphon $W$ at the known latent point pairs to the values of the probability matrices $\bbT^{(k)}$, and the regularization term $g(\cdot)$ in the objective imposes a prior on the overall graphon structure.
For example, we may apply a thin-plate spline term \cite{duchon1976interpolation} to estimate a smooth graphon assuming that points $\zeta^{(k)}$ are relatively evenly spaced throughout the interval $[0,1]$.

The assumptions in \eqref{equ_netinf_graphon} are weaker than those in \eqref{equ_netinf_probmat}, thus a wider range of applications are available.
In the example of estimating brain functional networks, functional connectivity of the same subject may be inferred for different sets of brain regions.
Additionally, climate network inference is often based on correlation or mutual inference measures, which decreases with geographic distance \cite{donges2009complex}.
Thus, separating inference of climate networks into multiple networks of subregions and applying \eqref{equ_netinf_graphon} may be more practical than estimating a single climate network for a large region, as connectivity is expected to be very low for far apart geographical locations.

\subsection{Examples for Network Inference Methods}
\label{Ss:combining_methods}

Up to this point, we have been considering a generic network inference problem in \eqref{equ_netinf}.
Both the formulations in \eqref{equ_netinf_probmat} and \eqref{equ_netinf_graphon} are applicable to existing network inference methods through specific choices of functions $f(\bbS,\bbX)$ and $L(\bbS)$. 

Consider examples for the function $f(\bbS,\bbX)$ that relate observed graph signals to the structure of the graphs.
Graph signals may be assumed to be smooth over their respective graphs \cite{kalofolias2016learn}, and we apply the penalty
\begin{equation}\label{equ_ref_f1}
f(\bbS,\bbX) = \sum_{k=1}^K \|\bbS^{(k)}\circ\bbZ^{(k)}\|_1,
\end{equation}
where $\bbZ^{(k)}_{ij} = \|\bbX^{(k)}_i-\bbX^{(k)}_j\|^2$ as in \cite{kalofolias2016learn}.
Alternatively, we may have graph signals that are the diffusion of noise through a graph filter \cite{segarra2017network,mateos2019connecting}.
In this case, we have stationary graph signals, where the signal covariance $\bbC$ commutes with the adjacency matrix $\bbS$. 
Defining sample covariance matrices as $\bbC^{(k)}=\frac{1}{R^{(k)}}\bbX^{(k)}(\bbX^{(k)})^\top$, we can write
\begin{equation}\label{equ_ref_f2}
f(\bbS,\bbX) = \sum_{k=1}^K \|\bbS^{(k)}\bbC^{(k)} - \bbC^{(k)}\bbS^{(k)}\|_F^2.
\end{equation}

In many applications, the graphs of interest are sparse, so it is common to apply a sparsity constraint for each graph \cite{friedman2008sparse,segarra2017network}.
We may apply this with the penalty function $L(\bbS)$ as
\begin{equation}\label{equ_ref_L1}
L(\bbS) = \sum_{k=1}^K \| \mathrm{vec}(\bbS^{(k)})\|_1.
\end{equation}
If, instead of separately inferring each graph, we wish to promote similar sparsity patterns, we may encourage edge similarity between graphs \cite{navarro2020joint,danaher2014joint} as 
\begin{equation}\label{equ_ref_L2}
L(\bbS) = \sum_{k<k'} \|\mathrm{vec}( \bbS^{(k)}-\bbS^{(k')} )\|_1,
\end{equation}
which requires graphs that are not only the same size, but are also on the same node set. 
Thus, the regularizer in~\eqref{equ_ref_L2} is applicable to our formulation in~\eqref{equ_netinf_probmat} but not to the one in~\eqref{equ_netinf_graphon}.

Combinations of the described examples for $f(\bbS,\bbX)$ and $L(\bbS)$ are common in existing works.
For instance, graph signal stationarity in \eqref{equ_ref_f2} and sparsity penalties for each graph via \eqref{equ_ref_L1} are applied in \cite{segarra2017network}.
Moreover, joint inference is performed in \cite{navarro2020joint} by combining \eqref{equ_ref_f2} and \eqref{equ_ref_L2}.

\section{Numerical Experiments}
\urlstyle{same}

We compare the performance of network topology inference methods with and without the augmentations in \eqref{equ_netinf_probmat} and \eqref{equ_netinf_graphon}, denoted by "Mod. 1" and "Mod. 2", respectively.
For all experiments, we apply the same signal model assumption $f(\bbS,\bbX)$ as \eqref{equ_ref_f2}, and we compare separate network inference via sparsity penalties \eqref{equ_ref_L1} and joint network inference via pairwise difference penalties \eqref{equ_ref_L2}.
For synthetic experiments, we sample from the graphon $W(x,y)=\frac{1}{2}(x^2+y^2)$.
The error of estimator $\hat{\bbS}$ is calculated as $\|\bbS-\hat{\bbS}\|_F/\|\bbS\|_F$, where the true GSO is given by $\bbS$.
Optimization problems are solved via the alternating direction method of multipliers (ADMM) \cite{boyd2011distributed}.\footnote{Implementations of our method are available at {\url{https://github.com/mn51/jointinf_graphs_graphon}}.}

\vspace{.1in}
\noindent{\bf Same node sets.}
We consider the case where all graphs are sampled from the same points within the graphon space, $\zeta^{(k)}=\zeta$. 
We estimate $K=3$ graphs with $N=30$ nodes as we observe an increasing number of signals for sample covariance computation.
We present in Fig.~\ref{fig_exp}a the comparison of separate and joint network inference methods with the augmentations in \eqref{equ_netinf_probmat} and \eqref{equ_netinf_graphon}, and without either. 
In both methods, the augmented formulations improve estimation performance significantly. 
The pairwise joint penalty \eqref{equ_ref_L2} enjoys the greatest improvement, as graphs not only possess node alignment required by \eqref{equ_ref_L2}, but they also follow our graph model assumption.

\vspace{.1in}
\noindent{\bf Node sets of different sizes.}
We consider the challenging case where the graphs have different latent point sets of different sizes $N^{(k)}\neq N^{(k')}$.
Unlike the previous experiment, we cannot apply \eqref{equ_netinf_probmat} or \eqref{equ_ref_L2}, so we consider only \eqref{equ_ref_f2} and \eqref{equ_ref_L1} with and without the joint graphon estimation from \eqref{equ_netinf_graphon}.
We consider $K=3$ graphs for node sets of $N=10,30,50$ and $N=15,30,45$.
For both cases, application of joint graphon inference results in consistent improvement, with increasing performance gap for larger number of observed signals.

\vspace{.1in}
\noindent{\bf Senate networks.}
Finally, we performed different-sized graph estimation with real-world data of U.S. congress roll-call votes \cite{lewis2020voteview}, and we set up senate vote signals as in \cite{navarro2020joint}.
We observe the $724$, $919$, and $612$ votes of congresses 103, 104, and 105, respectively, and we let the underlying true networks be obtained by separate estimation of each network using all available votes.
The number of nodes $N=101$ corresponds to the number of voters (100 senators and 1 President).
We estimate induced subgraphs of size $N=15,30,45$, where we only observe votes of senators corresponding to these subsets of nodes. 
In Fig.~\ref{fig_exp}c we observe that joint network and graphon estimation consistently outperforms separate inference, even though the true networks were estimated separately.
This demonstrates that the versatile nonparametric nature of graphons can aid the recovery of real-world graphs, which have not been explicitly drawn from a graphon model in the first place.

\section{Conclusion and Future Work}

We demonstrated a method to jointly estimate multiple networks under the assumption that they are sampled from the same graphon. 
To the best of our knowledge, this is the first method that leverages graphons to solve the challenging problem of inferring graphs of different sizes.
We demonstrated that our proposed maximum-likelihood-based method improves network estimation in synthetic and real-world experiments.
In terms of future directions, we plan to consider:
i)~The more challenging case where only noisy or partial information about the latent variables $\zeta^{(k)}$ is available, and
ii)~Other random graph models (beyond graphons) that can also generate graphs of different size while promoting different graph structural characteristics.

\bibliographystyle{ieeetr}
\bibliography{citations}

\end{document}